%% file: main.tex
\documentclass[10pt,twocolumn,letterpaper]{article}

\usepackage{cvpr}              

\input{preamble}

\definecolor{cvprblue}{rgb}{0.21,0.49,0.74}
\usepackage[pagebackref,breaklinks,colorlinks,allcolors=cvprblue]{hyperref}

\def\paperID{14970} 
\def\confName{CVPR}
\def\confYear{2025}

\title{VoteFlow: Enforcing Local Rigidity in Self-Supervised Scene Flow}

\author{
Yancong Lin$^{*,1,2}$, Shiming Wang$^{*,1}$, Liangliang Nan$^1$, Julian Kooij$^1$ and Holger Caesar$^1$ \\
$^1$TU Delft \quad  $^2$ETH Zurich\\
}

\input{macros}

\begin{document}

\maketitle

\def\thefootnote{*}\footnotetext{Equal contribution. Author ordering determined by coin flip.} 

\input{sec/0_abstract}    
\input{sec/1_introduction}
\input{sec/2_related_work}
\input{sec/3_methodology}
\input{sec/4_experiments}

\input{sec/5_conclusions}
\input{sec/6_acknowledgement}
{
    \small
    \bibliographystyle{ieeenat_fullname}
    \bibliography{main}
}


\end{document}




\input{sec/X_suppl}

{
    \small
    \bibliographystyle{ieeenat_fullname}
    \bibliography{main}
}


%% file: preamble.tex
\newcommand{\red}[1]{{\color{red}#1}}
\newcommand{\blue}[1]{{\color{blue}#1}}
\newcommand{\green}[1]{{\color{PineGreen}#1}}


%% file: macros.tex
\iftrue 
    \newcommand{\holger}[1]{\noindent}
    \newcommand{\ww}[1]{\noindent}
    \newcommand{\lin}[1]{\noindent}
\else
    \newcommand{\holger}[1]{[{\bf \color{orange} HC: #1}]}
    
    \newcommand{\lin}[1]{[{\bf \color{blue} YL: #1}]}

\fi

\newcommand{\fig}[1]{Fig.~\ref{fig:#1}}
\newcommand{\tab}[1]{Tab.~\ref{tab:#1}}

\usepackage{multirow}
\usepackage{arydshln}
\usepackage{siunitx}
\usepackage{xcolor}
\usepackage{graphicx}
\usepackage{amsmath}
\usepackage{mathtools}  
\usepackage{amsfonts} 
\usepackage{amssymb}
\usepackage{bbm}
\usepackage{dsfont}
\usepackage{booktabs}
\usepackage{comment}
\usepackage{float} 
\usepackage{dblfloatfix} 
\usepackage{pifont}
\usepackage{tikz}
\usepackage{relsize}
\usepackage{MnSymbol,bbding,pifont} 

\usepackage{xspace}


%% file: sec/0_abstract.tex
\begin{abstract}
Scene flow estimation aims to recover per-point motion from two adjacent LiDAR scans. However, in real-world applications such as autonomous driving, points rarely move independently of others, especially for nearby points belonging to the same object, which often share the same motion. Incorporating this locally rigid motion constraint has been a key challenge in self-supervised scene flow estimation, which is often addressed by post-processing or appending extra regularization. While these approaches are able to improve the rigidity of predicted flows, they lack an architectural inductive bias for local rigidity within the model structure, leading to suboptimal learning efficiency and inferior performance. 
In contrast, we enforce local rigidity with a lightweight add-on module in neural network design, enabling end-to-end learning.
We design a discretized voting space that accommodates all possible translations and then identify the one shared by nearby points by differentiable voting.  
Additionally, to ensure computational efficiency, we operate on pillars rather than points and learn representative features for voting per pillar.
We plug the Voting Module into popular model designs and evaluate its benefit on Argoverse 2 and Waymo datasets. 
We outperform baseline works with only marginal compute overhead. 
Code is available at \url{https://github.com/tudelft-iv/VoteFlow}. 
\end{abstract}

%% file: sec/1_introduction.tex
\section{Introduction}
\label{sec:intro}

\begin{figure}[tbp]
    \centering
    \includegraphics[width=0.45\textwidth]{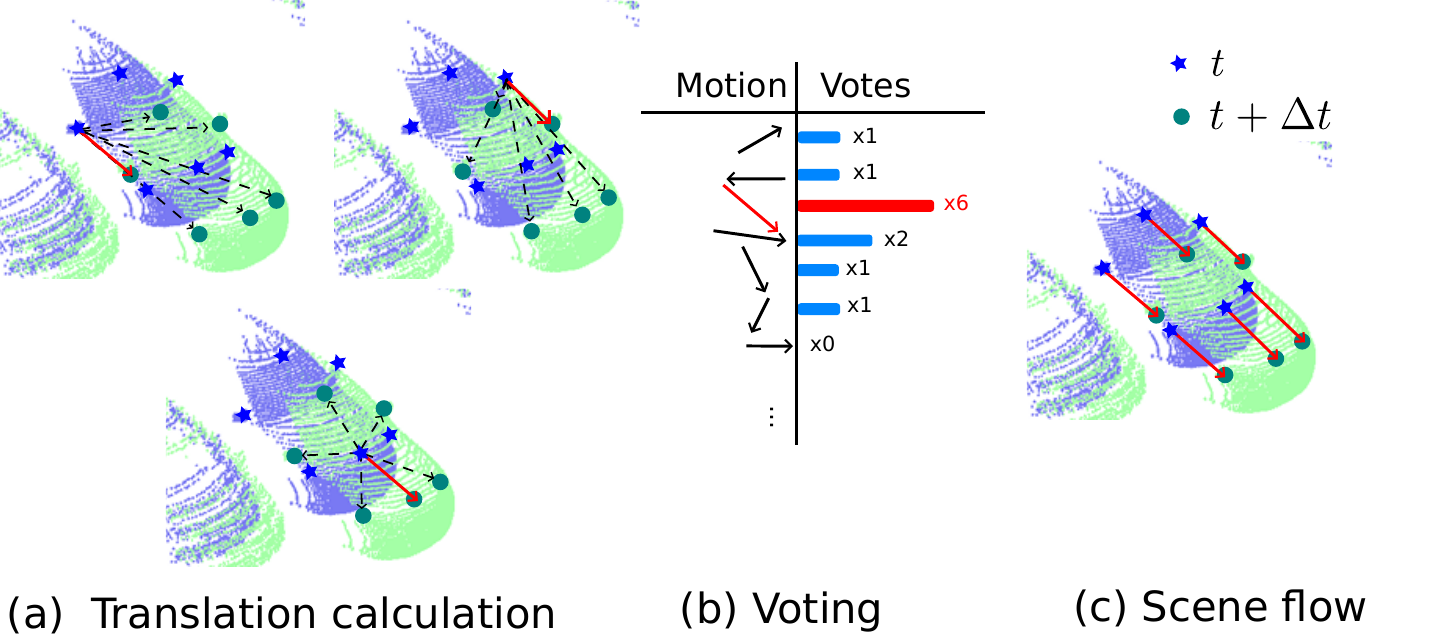}
    \caption{\textbf{Voting for identifying shared motion.} We take inspiration from motion rigidity and identify a motion shared by the majority via voting. We design a discrete voting space that encapsulates all possible translations that might occur within time $\Delta t$. For a given star \blue{$\bigstar$} at time $t$, we calculate the displacements between itself and its neighbors \green{\LARGE $\bullet$} at time $t + \Delta t$ and cast votes, defined by the cosine feature similarity between \blue{$\bigstar$} and \green{\LARGE $\bullet$}, to corresponding bins in the voting space. We accumulate votes from multiple spatially nearby points \blue{$\bigstar$} at time $t$. The voting result indicates the likelihood of a motion shared by nearby points. 
    Our differentiable voting is a light-weight add-on module compatible with popular model designs in scene flow estimation. In practice, voting takes pillars as input rather than individual points, thus reducing the computation substantially.
    }
    \label{fig:intro}
\end{figure}

Motion perception is essential for autonomous vehicles operating in dynamic environments. 
A crucial task in this domain, known as scene flow estimation, involves detecting per-point motion across consecutive LiDAR scans collected within short temporal intervals, e.g., 0.1 seconds given a 10Hz LiDAR scanner~\cite{dewan2016rigid,ushani2017learning,behl2019pointflownet, liu2019flownet3d}. 
Scene flow estimation has been the cornerstone in self-supervised scene understanding, which offers a way to interpret dynamic scenes without relying on extensively labeled data~\cite{zhai2020flowmot,  najibi2022motion, erccelik20223d, wang20224d, lentsch2024union}. 
For example, \cite{lentsch2024union} uses scene flow to associate moving objects over time and calculate object bounding boxes without any annotation cost. 
The generated object proposals provide valuable pseudo labels for subsequent model training. 
This strategy, also used in~\cite{najibi2022motion, wang20224d,dao2024label}, leverages the low-cost nature of data collection to scale efficiently and thus generates large quantities of pseudo labels.
Therefore, there is a perceivable demand for a robust scene flow estimator in autonomous driving.  

Scene flow methods typically assume \textit{motion rigidity}, i.e. nearby points on rigid objects share the same motion.
State-of-the-art methods exploit this assumption for self-supervised training through extra loss functions or regularizers~\cite{vidanapathirana2024multi, zhang2025seflow}, or hard-coded post-processing~\cite{chodosh2023re}. However, these models lack the ability to encode motion rigidity by design.
ICP-Flow\footnote{Winner of the Argoverse Scene Flow Challenge (unsupervised track) at CVPR 2024.}~\cite{lin2024icpflow} enforces motion rigidity by using Iterative Closest Point (ICP) to align pre-clustered points. However, this can produce substantial errors if the given clusters are over- or under-segmented (e.g., one cluster containing multiple close-by objects).

To overcome these limitations, we propose \textit{VoteFlow}, a novel self-supervised scene flow estimation method that integrates feature learning and motion rigidity as an architectural inductive bias within its network design.
The model extends the design of~\cite{jund2021scalable, vedder2023zeroflow, zhang2024deflow, zhang2025seflow}, which has good generalization and inference speed,
with a new differentiable Voting Module that identifies shared motion among neighboring points.
In particular, the module efficiently identifies translation-dominated motion, which often characterizes object motion over short intervals in autonomous driving, by locally collecting votes among nearby LiDAR features across possible translational directions, as shown in \fig{intro}. 
Our experiments show that VoteFlow outperforms the state of the art across most metrics not only on the Argoverse 2 benchmark but also on the Waymo Open dataset \textit{without} retraining for that dataset.
To summarize, our contributions are as follows:
\begin{itemize}
\item We introduce VoteFlow - a self-supervised scene flow estimator that encodes locally rigid motion by design. 

\item The core design of VoteFlow is a differentiable Voting Module - a light-weight add-on module that enables translation voting and end-to-end feature learning.

\item VoteFlow outperforms state-of-the-art methods on the Argoverse 2 dataset with a considerably low inference latency.
It also excels on the Waymo Open dataset in cross-dataset tests, outperforming baselines optimized on this particular dataset. 
\end{itemize}

%% file: sec/2_related_work.tex
\section{Related work}

\label{sec:related_work}
In this section we discuss prior works on scene flow estimation, approaches to enforce motion rigidity, and other usages of voting schemes in point cloud processing.

\subsection{Scene flow estimation}
Numerous works have emerged in scene flow estimation from large-scale point clouds in autonomous driving scenarios. 
Early works on scene flow, such as~\cite{behl2019pointflownet,liu2019flownet3d, huang2022dynamic,jin2022deformation}, are supervised mainly by learning from annotated data, either from fully labeled real-world datasets~\cite{geiger2013vision, caesar2020nuscenes, waymo} or synthetic datasets~\cite{Dosovitskiy_2015_ICCV}. 
However, annotating large-scale datasets is costly, thus demanding the need for unsupervised alternatives.
Later works remedy the need for labels by exploiting cycle consistency~\cite{baur2021slim,mittal2020just, wu2020pointpwc} between the forward and backward flows. 
Recently, test-time optimization techniques have been prevalent~\cite{li2021neural, li2023fast, chodosh2023re}.
This family of works builds on top of multi-layer perceptrons (MLPs) and optimizes the cycle loss during inference.
While the accuracy of these methods is outstanding, the inference time remains extended (up to several minutes~\cite{li2021neural, chodosh2023re}) due to repeated optimization passes, making these methods impractical for online applications.
To remedy the costly test-time inference, \cite{vedder2023zeroflow, zhang2025seflow} propose feed-forward neural networks as an alternative to test-time optimization, allowing for better generalization and faster inference. Particularly, \cite{vedder2023zeroflow} is capable of real-time inference. 
Our work also employs fast feed-forward networks.

\subsection{Motion rigidity in scene flow estimation}
\label{sec:rigid-flow}
A common assumption of scene flow estimation methods is that objects exhibit non-deformable motion.
This rigid motion assumption can be exploited in several ways.
One approach is to explicitly identify individual rigid objects and estimate a single rigid transformation for each such object.
For instance, ICP-Flow~\cite{lin2024icpflow} clusters points into objects during a pre-processing step to enforce per-cluster rigid motion.
Clustering can also be applied in post-processing~\cite{chodosh2023re} to enforce rigid motion.
However, clustering can produce substantial errors if objects are over- or under-segmented (e.g., one cluster containing multiple close-by objects),
and clustering parameters are not end-to-end optimizable and therefore require extensive parameter tuning.
A similar strategy has also been adopted in RigidFlow~\cite{li2022rigidflow}.
Self-supervised training can also enforce consistent motion in a local neighborhood through extra loss functions or regularizers~\cite{vidanapathirana2024multi, zhang2025seflow}.
However, such models rely on training the network to incorporate the rigidity assumption in the network weights, which is less data-efficient and harder to train due to potentially conflicting loss terms.
PointPWC~\cite{wu2020pointpwc} improves locality and rigidity by constructing cost volumes and aggregating features from nearby points. However, its design is resource-intensive and does not explicitly enforce rigidity. 
Our model differs from previous works by explicitly encoding motion rigidity as an architectural inductive bias in the network design.

\subsection{Voting in point cloud processing}
Hough Voting~\cite{duda1972use} is a classic image processing technique initially designed for extracting geometric primitives from images, such as lines and circles. Later Hough Voting has been extended to find arbitrary shapes~\cite{ballard1981generalizing, lin2021nerd++, lin2022vpd}. 
Hough Voting also has been widely used in processing point clouds, for segmentation~\cite{milletari2015robust}, detection~\cite{lehmann2011fast}, tracking~\cite{milletari2015universal} and 6D pose estimation~\cite{sun2010depth}. 
ICP-Flow~\cite{lin2024icpflow} is a highly relevant work that extends the usage of majority voting to scene flow, where voting is used to localize the most dominant translation and to initialize the Iterative Closest Point algorithm.
Our work also leverages majority voting for scene flow estimation but uses learned rather than hand-crafted features.
Recently, there has been a line of works that encodes Hough Voting in deep learning, thus allowing for voting by learned features~\cite{lee2021deep, qi2019deep,lin2020ht,
velizhev2012implicit, woodford2014demisting}. 
Our work shares the same spirit in combining voting with learned features.

%% file: sec/3_methodology.tex
\section{Methodology}
\label{sec:methodology}
This section describes our proposed method, VoteFlow,
and its novel Voting Module for efficiently identifying shared translations across local regions in LiDAR feature maps.

\begin{figure*}[]
    \centering
    \includegraphics[width=\textwidth]{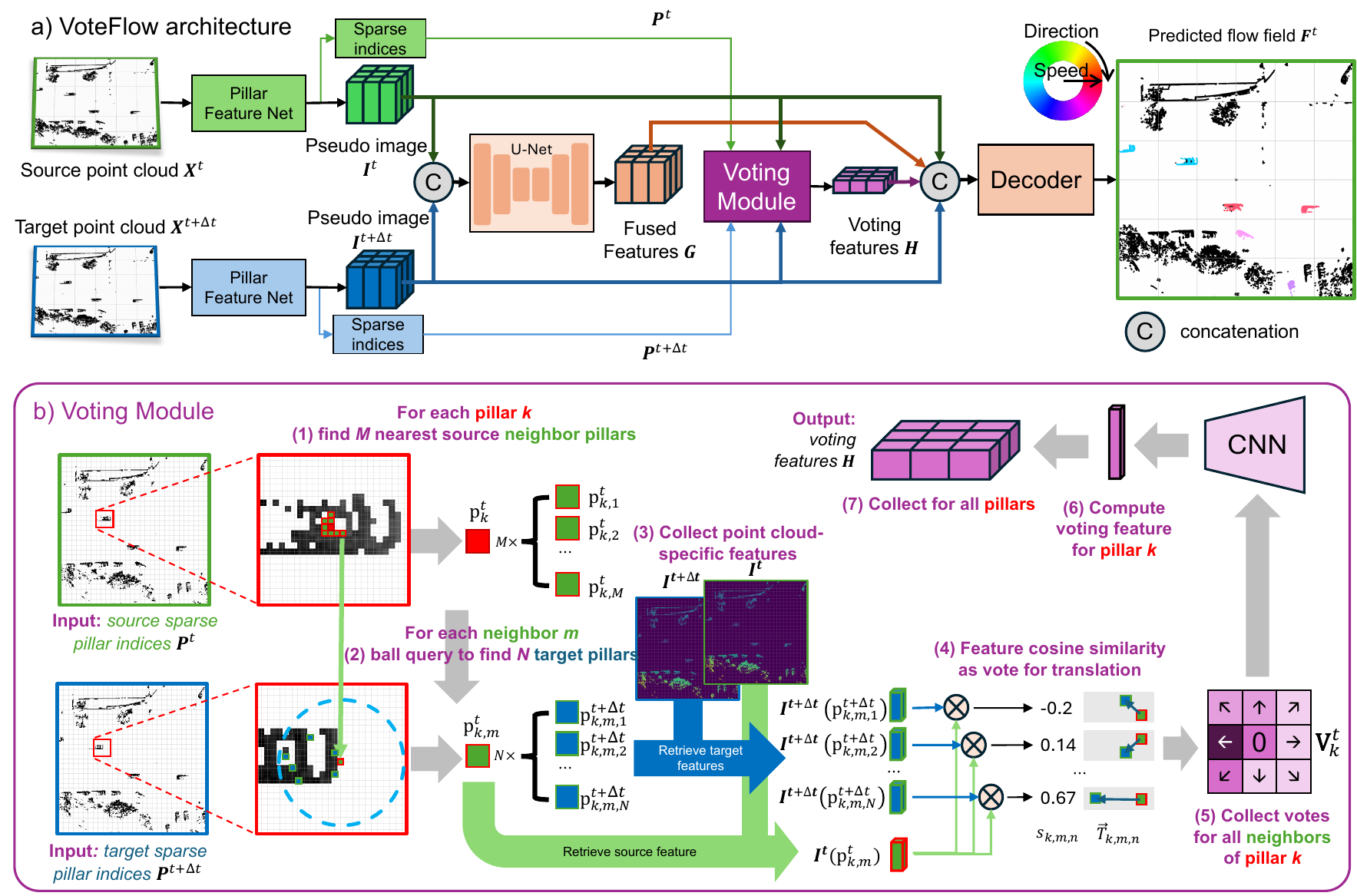}
    \caption{\textbf{The overall architecture of VoteFlow.}
    VoteFlow introduces a new end-to-end optimizable Voting Module that matches features in the local region around each pillar. Such matches are used to vote for a translation of the pillar. A CNN module summarizes the resulting votes as a `voting feature'. The voting features are the output of the module and passed on to the decoder.}
    \label{fig:overall}
\end{figure*}

\subsection{Problem statement}
Scene flow estimation aims to recover a flow field (or equivalently point-wise translations) $\textbf{F}^{t} \in {\mathbb{R}^{3\times L}} = \{ \textbf{f}_{l} \in \mathbb{R}^{3} \}_{l=1}^{L} $ from a pair of consecutive LiDAR scans $\textbf{X}^{t}$ and $\textbf{X}^{t+\Delta t}$, captured by an autonomous vehicle at time $t$ and $t+\Delta t$, such that $\textbf{X}^{t} + \textbf{F}^{t}  \approx \textbf{X}^{t+\Delta t}$. $\textbf{X} \in {\mathbb{R}^{3\times L}} = 
\{ \textbf{x}_{l} \in \mathbb{R}^{3} \}_{l=1}^{L} $ denotes a point cloud with $L$ points. 
It is worth noting that $\textbf{X}^{t}$ and $\textbf{F}^{t}$ are of the same size, while $\textbf{X}^{t}$ and $\textbf{X}^{t+\Delta t}$ may differ in size.
Following common practice in scene flow estimation~\cite{Argoverse2, vedder2023zeroflow, zhang2025seflow}, we also assume the ego motion is available, \textit{i.e.,} the input points $\textbf{X}^{t}$ and $\textbf{X}^{t+\Delta t}$ have already been compensated by ego motion.
Our goal is to develop a neural network that takes $\textbf{X}^{t}$ and $\textbf{X}^{t+\Delta t}$ as input and predicts $\textbf{F}^{t}$, without relying on supervision.

\subsection{Overview of VoteFlow}
\fig{overall} shows an overview of VoteFlow. Given two consecutive LiDAR scans as input, the model first applies a Pillar Feature Net~\cite{lang2019pointpillars} (``pillarization'') to convert both scans into a bird-eye-view pseudo image, where each grid cell (``pillar'') represents a 2D location around the ego-vehicle and has an associated embedding.
The model concatenates both pseudo images and learns features via a U-Net~\cite{ronneberger2015unet} backbone.
Subsequently, the learned features go through our novel Voting Module, which creates a voting space for each \textit{non-empty} pillar and applies convolution within the voting space.
Later, the model retrieves points-wise features from the pseudo images, the fused features, and the voting features, appended by per-point offset with respect to the pillar center.
The decoder converts point-wise features into point-wise scene flow.

We note that pillar representations of point clouds in autonomous driving are often sparse.
A statistical analysis of the Argoverse 2 dataset~\cite{Argoverse2} shows that more than 90\% percent of pillars are empty\footnote{As a common practice in scene flow estimation, ground points are removed from point clouds, as they provide little cue to predict motion~\cite{li2021neural,vedder2023zeroflow}, which increases the proportion of empty pillars substantially.}.
Our proposed voting scheme exploits the sparsity of these feature maps.

The following subsections elaborate on each component.

\subsubsection{Pillarization and backbone design}
The model follows previous work~\cite{jund2021scalable, vedder2023zeroflow} and adopts the same setup during pillarization and backbone feature extraction. To be specific, we set the pillar size to be $\delta_{y} \times \delta_{x}$, resulting in a pseudo image of spatial size $H \times W $ for both LiDAR scans $\textbf{X}^{t}$ and $\textbf{X}^{t+\Delta t}$. The backbone takes as input the concatenation of both pseudo images $\textbf{I}^{t}$ and $\textbf{I}^{t+\Delta t}$ along the channel dimension and generates a fused feature map $\textbf{G}$ of spatial size $H \times W$ using a U-Net with skip connections.

\subsubsection{Voting Module}
To exploit motion rigidity in the network design, 
we observe that neighboring pillars on the same object are expected to share the same translation.
Thus, we seek to identify the dominant translation from a number of nearby pillars.
Our novel Voting Module creates a discretized voting space $\textbf{V}$ for each non-empty pillar at time $t$ that covers all possible translations that may occur within $\Delta t$.
It then identifies $M$ neighboring pillars at time $t$,
and each of the $M$ neighboring pillars casts votes for possible translations in $\textbf{V}$ by identifying all neighboring pillars within a predefined radius at time $t + \Delta t$.
After aggregating evidence on consistent motion by all neighboring pillars,
a higher vote in $\textbf{V}$ indicates more evidence for a particular translation.
We elaborate on this concept as follows.

The module takes as input
a set of indices of non-empty \textit{source} pillars $\textbf{P}^{t}$ at time $t$, 
the non-empty \textit{target} pillar indices
$\textbf{P}^{t+\Delta t}$ at the next time instance,
and the pseudo images ( $\textbf{I}^{t}$ and $\textbf{I}^{t+\Delta t}$) from the Pillar Feature Net.
It will output $\textbf{H}$, which contains a  \textit{voting feature} for each pillar in $\textbf{P}^{t}$.

For each pillar $\textbf{p}_{k}^{t} \in \textbf{P}^{t}$ the module selects the $M$ spatially closest pillars $\{ \textbf{p}_{k,m}^{t} \}_{m=1}^{M}$ from the same time step (this set includes the pillar $k$ itself).
For each neighbor pillar $\textbf{p}_{k,m}^{t}$ at time $t$ a set of 
$N$ spatially nearby pillars $\{ \textbf{p}_{k,m,n}^{t+\Delta t} \}_{n=1}^{N}$ at time $t+\Delta t$ has been selected from the available pillar indices $\textbf{P}^{t+\Delta t}$ using the \textit{ball query} function~\cite{ravi2020pytorch3d}.

Note that each pair $(\textbf{p}_{k,m}^{t}, \textbf{p}_{k,m,n}^{t+\Delta t})$ represents a possible (discretized) 2D translation $\overrightarrow{T}_{k,m,n}$ of the $m$-th neighborhood pillar of $k$.
We use the cosine similarity of their corresponding pseudo images $\textbf{I}^{t}(\textbf{p}_{k,m}^{t})$ and $\textbf{I}^{t + \Delta t}(\textbf{p}_{k,m,n}^{t+\Delta t})$,
as a `voting score' $s_{k,m,n} \in [-1, +1]$ for the translation $\overrightarrow{T}_{k,m,n}$,
such that a high score provides evidence in favor of this translation.
Next, an empty voting space $\textbf{V}_{k}^{t}$ is initialized for pillar $k$.
With the range of allowed translation within $\Delta t$ seconds set to $(x_{min}, x_{max})$ and $(y_{min}, y_{max})$, 
$\textbf{V}_{k}^{t}$ is simply a discrete grid of size $H_{v} \times W_{v}$, where $H_{v}= \frac{y_{max}-y_{min}}{\delta_{y}}$ and $W_{v}= \frac{x_{max}-x_{min}}{\delta_{x}}$.
All $M \times N$ votes are collected by summing their scores in the correct bins,
\begin{equation}
\textbf{V}_{k}^{t}(\overrightarrow{T}_{k,m,n}) \leftarrow \textbf{V}_{k}^{t}(\overrightarrow{T}_{k,m,n}) + s_{k,m,n} \quad \forall m, n.
\end{equation}

Finally, all votes can be summarized.
Simply taking the \textit{argmax} on $\textbf{V}_{k}^{t}$ has several downsides, however.
It would only identify a single dominant direction, which can be erroneous during early training,
and it could only produce a coarse discretized motion.
Instead, our Voting Module summarizes the vote as a continuous feature vector such that the later decoder can access all voting information.
This is achieved by applying two convolutional layers (with ReLUs) to each $\textbf{V}_{k}^{t}$ and flattening the result.
The voting features for all non-empty pillars are collected as $\textbf{H}$
(the decoder will not need to lookup features for empty pillars).

\subsubsection{Decoder}
The decoder transforms the learned features into flow predictions.
Similar to baseline works~\cite{vedder2023zeroflow, jund2021scalable}, the decoder retrieves point-wise features from pseudo images $\textbf{I}^{t}$ and $\textbf{I}^{t + \Delta t}$, the fused features $\textbf{G}$, but now also the voting features $\textbf{H}$ by reusing the point-to-pillar indices.
Additionally, the decoder also appends point-wise offsets (with respect to the pillar center) to point-wise features. 
The point-wise features go through 4 layers of fully connected layers (with ReLUs) and become per-point scene flow prediction. 
Notably, the decoder differs from SeFlow~\cite{zhang2025seflow}, as it uses fully connected layers rather than GRU layers. This is to reduce the computational cost during both training and inference.

\subsection{Loss functions}
We train VoteFlow in a self-supervised manner,
adopting the loss functions from SeFlow~\cite{zhang2025seflow}.
The first loss is the common bidirectional Chamfer loss~\cite{li2021neural} $\mathcal{L}_{chamfer}$ that minimizes the distance between $\textbf{\^{X}}^{t}$ and $\textbf{\^{X}}^{t+\Delta t}$, where $\textbf{\^{X}}^{t} = \textbf{X}^{t} + \textbf{F}^{t}$.
Let $ \mathcal{D}(\textbf{x}, \textbf{Y}) = \min\limits_{\textbf{y}_{k} \in \textbf{Y}}\| \textbf{x} - \textbf{y}_{k} \|$,
then
\begin{align*}
\mathcal{L}_{chamfer} = \frac{1}{\vert \textbf{\^{X}}^{t} \vert}\sum_{\textbf{x} \in \textbf{\^{X}}^{t}}^{} \mathcal{D}(\textbf{x}, \textbf{X}^{t+\Delta t}) \\ + 
\frac{1}{\vert \textbf{X}^{t+\Delta t} \vert}\sum_{\textbf{x} \in \textbf{X}^{t+\Delta t}}^{} \mathcal{D}(\textbf{x}, \textbf{\^{X}}^{t}).
\label{eq:loss}
\end{align*} 
The other loss functions are: 
$\mathcal{L}_{dynamic}$, 
$\mathcal{L}_{static}$, and 
$\mathcal{L}_{cluster}$.
$\mathcal{L}_{dynamic}$ is the same as $\mathcal{L}_{chamfer}$ but only applies to dynamic points, which are predefined by an offline method DUFOMap~\cite{daniel2024dufomap}. The purpose is to handle class imbalance as dynamic points are the minority of the points.
$\mathcal{L}_{static}$ is only imposed on static points, which encourages static points to have zero flows. 
$\mathcal{L}_{cluster}$ is only on pre-clustered points computed by HDBSCAN~\cite{mcinnes2017hdbscan} and encourages the points from the same cluster to have the same flow predictions.
The total loss function is defined as $\mathcal{L}_{total} = \mathcal{L}_{chamfer} + \mathcal{L}_{dynamic} + \mathcal{L}_{static} + \mathcal{L}_{cluster}$.
We refer the readers to baseline work~\cite{zhang2025seflow} for details.


\subsection{Implementation details}
We follow previous works~\cite{vedder2023zeroflow, zhang2025seflow} and Argoverse 2 official evaluation protocol to configure the hyperparameters.
All point clouds are cropped within $102.4 \times 102.4 $ meters and compensated by ego motion. Ground points have also been removed using rasterized HD maps.
We set pillar sizes $\delta_{x}$ and $\delta_{y}$ to be 0.2 meters and maximal translations $x_{max}$ and $y_{max}$ to be 2 meters, resulting in a $20\times 20$ voting space $\textbf{V}$ per pillar where $H_{v}$ and $W_{v}$ are equal to 20.
To conduct voting, we select $M=8$ 
nearest neighbors for a given pillar at $t$ and sample $N=128$ 
neighboring pillars at $ t+ \Delta t$. 
The same setup also applies to the Waymo Open Dataset.

During training, we train VoteFlow for 12 epochs using Adam~\cite{kingma2014adam} optimizer and set the learning rate to the $2\times e^{-4}$, which is later decreased by 10 after 6 epochs. 

%% file: sec/4_experiments.tex
\section{Experiments}
\label{sec:experiments}


\begin{table*}
\centering
\resizebox{\textwidth}{!}{
\begin{tabular}[width=\textwidth]{cccccccccccc}
\toprule
\multirow{3}{*}{Method}             & \multirow{3}{*}{Labels}   & \multicolumn{6}{c}{Bucketed Normalized EPE $\downarrow$}            & \multicolumn{4}{c}{3-way EPE $\downarrow$ (in meters)} \\ 
\cline{3-12}
                                    &                        & \multicolumn{5}{c}{Dynamic (normalized EPE)}                  & Static (EPE, in meters) & \multirow{2}{*}{Avg.} & \multirow{2}{*}{FD} & \multirow{2}{*}{BS} & \multirow{2}{*}{FS} \\
\cline{3-8}
                                    &                        & \multicolumn{1}{c}{Mean} & Car & O. V. & Pd. & W. V & Mean   &             &            &              &              \\ 
\cline{1-12}
FastFlow3D~\cite{jund2021scalable}      & $\checkmark$ &  0.532          & 0.243  & 0.391 & 0.982  &  0.514  & 0.018  &  0.062  &   0.156           &       0.005       &  0.024              \\ 
DeFlow~\cite{zhang2024deflow}          & $\checkmark$ &  0.276          & 0.113  & 0.228 & 0.496  &  0.266  & 0.022  &  0.034  &   0.073           &        0.004      &  0.025              \\ 
TrackFlow~\cite{khatri2025can}       & $\checkmark$ &  0.269          & 0.182  & 0.305 & 0.358  &  0.230  & 0.045  &  0.047  &   0.103           &        0.002      &  0.037              \\ 
Flow4D~\cite{kim2024flow4d}       & $\checkmark$  & 0.174          & 0.096  & 0.167 & 0.278  &  0.155  & 0.012 &   0.025      &       0.057     &  0.003 & 0.015   \\
\midrule
FastNSF~\cite{li2023fast}         &           &  0.383          & 0.269  & 0.413 &  0.500 &  0.325  & 0.074  & 0.112   &  0.163            &         0.091      &    0.081     \\ 
NSFP~\cite{li2021neural}            &         &  0.422          & 0.251  & 0.331 &  0.723 &  0.383  & 0.028  &  0.061  &  0.116      &        0.034    & 0.032      \\ 
ZeroFlow ~\cite{vedder2023zeroflow}        &             &  0.594          & 0.327  & 0.476 &  0.966 &  0.608  & 0.020  &   -      &  -                 &            -        &  -     \\ 
ICP-Flow~\cite{lin2024icpflow}         &              &  0.331          & \textbf{0.195}  & 0.331 &  0.435 &  0.363  &  0.027 &   0.065 & 0.137 &    0.025   & 0.033 \\ 
SeFlow~\cite{zhang2025seflow}          &                &  0.309          & 0.214  & 0.292 &  0.463 &  0.267  &  \textbf{0.014} &  0.049  & 0.121   &  \textbf{0.006}  &  0.022 \\ 
VoteFlow (ours) & & \textbf{0.289} & 0.202 & \textbf{0.288} & \textbf{0.417} & \textbf{0.249} & \textbf{0.014} & \textbf{0.046} &  \textbf{0.114} & \textbf{0.006} & \textbf{0.018} \\
 
\bottomrule
\end{tabular}
}
\caption{\textbf{Comparison on Argoverse 2 \underline{test} split.} We compare all models using the Bucketed Normalized EPE, allowing for fine-grained analysis on individual classes, including Car, Other Vehicles (O. V.), Pedestrian (Pd), and Wheeled VRU (W. V.). The dynamic normalized EPE is a \textit{ratio} as the endpoint error has been normalized by speed, while the other EPE metrics are in meters~\cite{khatri2025can}. All results are from the Argoverse 2 Scene Flow Challenge Leaderboard. Our VoteFlow performs the best among all self-supervised models on mean dynamic normalized EPE. On individual classes, VoteFlow achieves the best result on Pedestrian and Wheeled VRU and performs competitively on Car and Other Vehicles, with only marginal gaps to the best. The inference time of VoteFlow is approximately 25.6 ms per sample on an A100 GPU.
}
\label{tab:exp_argo}
\end{table*}

\begin{table}[htbp]
\centering
\resizebox{\columnwidth}{!}{
\begin{tabular}[width=0.45\textwidth]{c c c ccc}
\toprule
\multirow{2}{*}{Method}                                     & \multirow{2}{*}{Labels} & Same-domain  &  \multicolumn{3}{c}{EPE $\downarrow$ (in meters)}                  \\ 
\cline{4-6}
                                                            &                           & training & FD & FS & BS            \\ 
\cline{1-6}
FastFlow3D~\cite{jund2021scalable}      & $\checkmark$                                            & $\checkmark$ & 0.195 & 0.025 & 0.015\\ 
DeFlow~\cite{zhang2024deflow}         & $\checkmark$                                          & $\checkmark$ & 0.098 & 0.026 & 0.010\\ 
\midrule
FastNSF~\cite{li2023fast}              &                           & $\checkmark$& 0.301 & 0.015 & 0.040   \\ 
NSFP~\cite{li2021neural}               &                         & $\checkmark$ & 0.171 & 0.108 & 0.022   \\ 
ZeroFlow ~\cite{vedder2023zeroflow}    &                           & $\checkmark$ & 0.216 & 0.015 & 0.024      \\ 

SeFlow~\cite{zhang2025seflow}          &                         & $\checkmark$ & 0.151 & 0.018 & \textbf{0.011}     \\ 
VoteFlow (ours) & & $\checkmark$ & \textbf{0.117} &  0.015 & 0.016 \\ 
\midrule
SeFlow~\cite{zhang2025seflow}         &         &  & 0.155 & 0.018 & 0.013 \\
VoteFlow (ours)      &          &  & 0.142 & \textbf{0.014} & 0.012            \\ 
\bottomrule
\end{tabular}
}
\caption{\textbf{Comparison on Waymo Open validation split.} We report the EPE results on foreground dynamic (FD), foreground static (FS), and background static (BS). Although \textit{without} training on Waymo, our model exhibits the best result across self-supervised models, indicating the generalization ability of VoteFlow across datasets. }
\label{tab:exp_waymo}
\end{table}


\begin{figure*}[htbp]
    \centering
    \includegraphics[width=\textwidth]{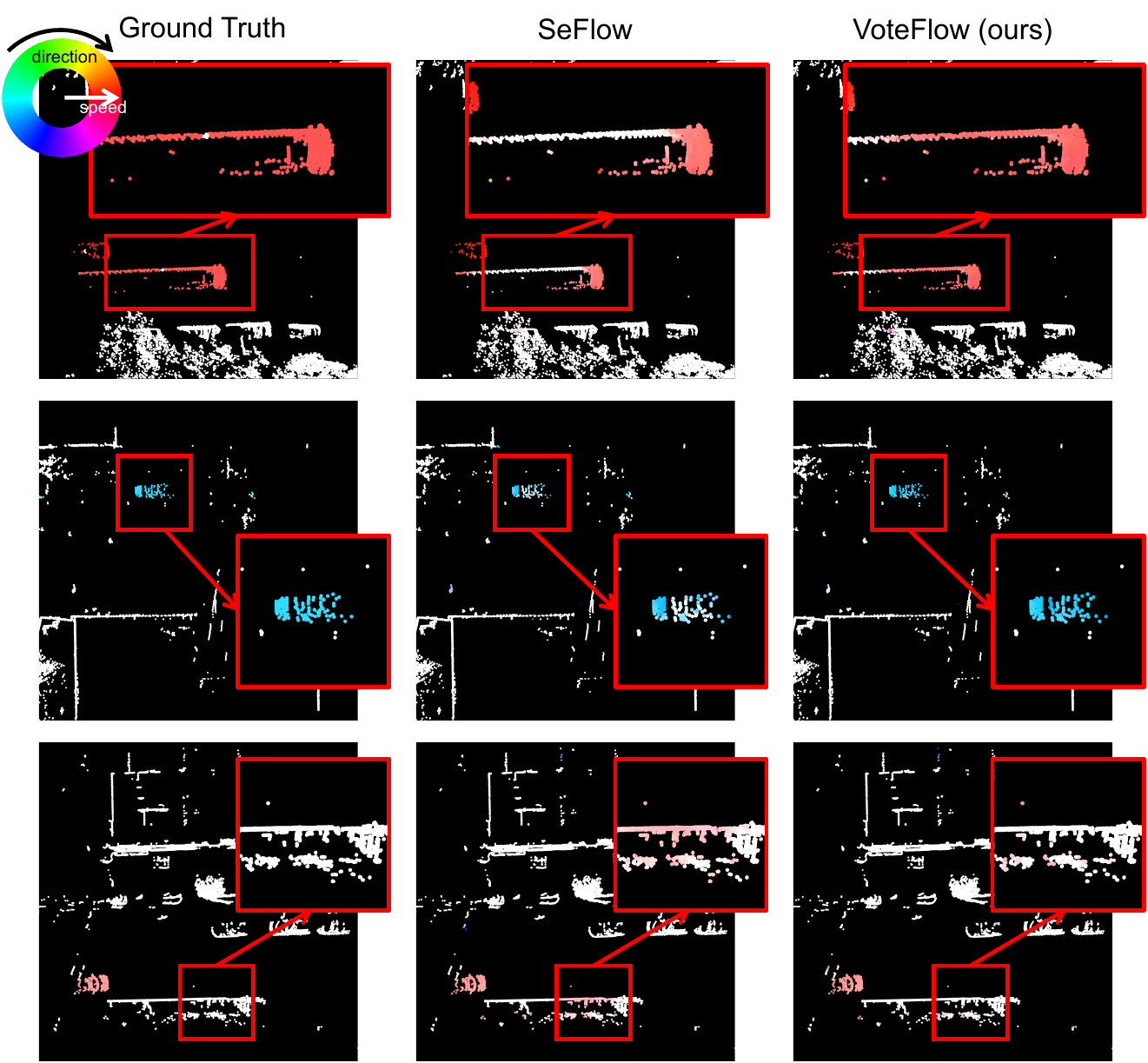}
    \caption{\textbf{Qualitative results on Argoverse 2 validation set.} 
    Colors indicate directions and saturation of the color indicates the scale of the flow estimation. 
    Thanks to the local rigidity prior, our VoteFlow predicts more consistent and coherent flow over objects compared to our baseline.
}
    \label{fig:qualitative_results}
\end{figure*}


We validate our approach on both the Argoverse 2~\cite{Argoverse2} and Waymo~\cite{waymo} datasets, which are the most commonly used datasets for scene flow in autonomous driving. 
This section describes the datasets, evaluation metrics and baselines, followed by evaluations and discussions.

\subsection{Datasets}

The \textbf{Argoverse 2} dataset~\cite{Argoverse2} contains LiDAR scans captured by two roof-mounted 32-beam LiDARs, with a 0.1 second interval between consecutive scans. All LiDAR scans are compensated by ego motion. The ground points within the dataset are removed according to a rasterized height map. We conduct evaluations on the official test split. 
The Argoverse 2 2024 Scene Flow Challenge~\cite{khatri2025can} has provided many baseline results on Argoverse 2.

We also evaluate on the \textbf{Waymo Open}~\cite{waymo} benchmark similar to~\cite{vedder2023zeroflow, zhang2025seflow}.
It contains 798 training and 202 validation scenes, with a LiDAR frequency of 10 Hz.
Ground points are removed like Argoverse 2, and LiDAR scans are provided with ego-motion compensation.

\subsection{Evaluation}
We adopt the Bucketed Normalized EPE metric for evaluation on Argoverse 2~\cite{khatri2025can}, which directly measures performance disparities across semantic classes and speed profiles, allowing us to normalize comparisons between classes moving at different speeds.

EPE stands for endpoint error between a predicted flow and ground truth flow.
There are four classes in total: Car, Pedestrian, Other Vehicles, and Wheeled VRU. For each class, we compute the static EPE (in meters), which is the average endpoint error of all static points (with a speed lower than $0.4 m/s$) and the dynamic normalized EPE (in ratio, defined as the mean of normalized endpoint errors over a set of predefined speed profiles, e.g., $0.4-0.8 m/s$, $0.8-1.2m/s$, etc.
Normalizing the endpoint error by speed ensures that high-speed objects (e.g., cars) and low-speed objects (e.g., pedestrians) are fairly evaluated.
For example, a $0.5 m/s$ error on a car moving $20 m/s$ is negligible ($<2.5$\%), while a $0.5 m/s$ error on a pedestrian moving $0.5 m/s$ fails to depict the pedestrian’s motion. 

On the Waymo Open dataset, we adopt the three-way EPE (in meters) evaluated on foreground static, foreground dynamic, and background, since Bucketed Normalized EPE is not yet available on the Waymo Open Dataset~\cite{khatri2025can}.

\subsection{Baselines}
We compare VoteFlow against five prominent baselines including NSFP~\cite{li2021neural}, FastNSF~\cite{li2023fast}, ZeroFlow~\cite{vedder2023zeroflow}, ICP-Flow~\cite{lin2024icpflow} and SeFlow~\cite{zhang2025seflow}.  
NSFP~\cite{li2021neural} and FastNSF~\cite{li2023fast} use test-time optimization. 
ZeroFlow~\cite{vedder2023zeroflow} uses pseudo labels produced by NSFP~\cite{li2021neural} for training and exhibits strong scalability when ample data is available.
ICP-Flow~\cite{lin2024icpflow} incorporates motion rigidity explicitly into the design. However, it lacks the ability to learn strong features from data. 
SeFlow~\cite{zhang2025seflow} is a recent work from ECCV'24 that achieves top performance on the leaderboard, thus providing a good reference for evaluating our model. 
%
Although VoteFlow focuses on self-supervised learning, we also include several fully supervised baselines, namely Flow4D~\cite{kim2024flow4d}, DeFlow~\cite{zhang2024deflow}, FastFlow3D~\cite{jund2021scalable} and TrackFlow~\cite{khatri2025can}.

\subsection{Comparison on Argoverse 2}
Tab.~\ref{tab:exp_argo} shows the comparison on the Argoverse 2 \underline{test} split, including both supervised and self-supervised approaches. We directly take the results from the Argoverse 2 Scene Flow Challenge Leaderboard\footnote{\url{https://www.argoverse.org/sceneflow.html}}.
We focus primarily on the dynamic errors over different categories, represented by normalized EPE (in ratio), since static errors, indicated by EPE (in meters) are negligible. 

On average, we improve over the previous best, SeFlow, by $2.0\%$pt (percentage points) in dynamic normalized EPE averaged over all four categories, indicating the effectiveness of the motion rigidity prior in VoteFlow.  
Among all four categories, VoteFlow outperforms SeFlow on Car, Other Vehicles, Pedestrian, and Wheeled VRU by a margin of $1.2\%$pt, $0.4\%$pt, $4.6\%$pt, and $1.8\%$pt, respectively. An illustrative example to demonstrate the performance gap is that, given a car moving at 20 m/s, we reduce the EPE error by approximately 2.4 cm, as the normalized EPE is equivalent to the estimated speed error from EPE divided by ground truth speed.
ICP-Flow is a strong baseline that achieves the best results on Car. However, its performance lags behind by a large margin on Wheeled VRU, thus leading to an inferior overall result.

Compared to supervised models, there is still a performance gap of $11.5\%$pt between our VoteFlow and the best-performing Flow4D. Notably, Flow4D differs from others in using multiple LiDAR consecutive scans during training, indicating that exploring temporal information is a highly effective approach to further enhance the performance of scene flow estimation.

\subsection{Comparison on Waymo Open}

We also compare various models on the Waymo Open dataset, as shown in \tab{exp_waymo}. The evaluation metric is EPE (in meters), evaluated on Foreground Dynamic (FD), Foreground Static (FS), and Background Static (BS). 
When training and valuation are both on Waymo, our model outperforms SeFlow by a margin of approximately 3 cm on FD.
Notably, our VoteFlow exhibits the best performance among all self-supervised models on FD and FS even if we load the pretrained checkpoint on the Argoverse 2 dataset and directly test its performance on Waymo. 
VoteFlow excels \textit{without} time-consuming and data-demanding training on the Waymo Open dataset, indicating its strong robustness across datasets.

\subsection{Analysis on voting}

\begin{figure}[tbp]
    \centering
    \includegraphics[width=\columnwidth]{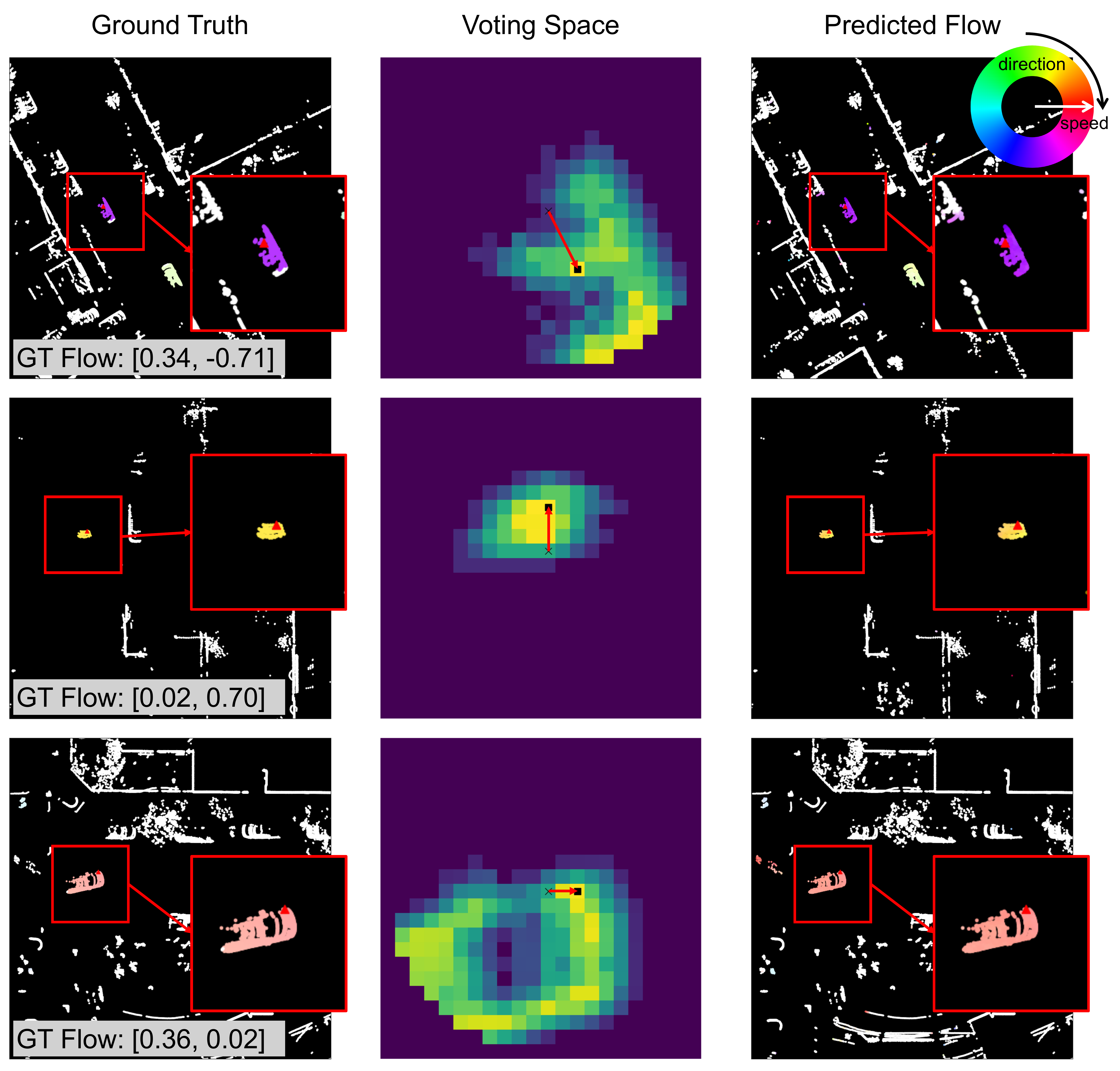}
    \caption{\textbf{Visualization of the voting space.}
    We select a pillar representing a moving object and plot its voting space, where the center indicates zero translations and boundaries indicate minimal and maximal translations along both dimensions. 
    As shown in the voting space, the red arrow aligns with the ground truth flow (up to quantization errors). Overall, the voting space depicts the heatmap of the object's motion. For example, in the top row, the object is expected to move towards the bottom right, and the predicted heat map also has high responses along the same direction.
    }
    \label{fig:voting_feats}
\end{figure}

We analyze the voting space qualitatively in \fig{voting_feats} to validate its role in scene flow prediction. The middle column shows the voting space for a given pillar, indicated by a red triangle marker (\red{$\blacktriangle$}) in the ground truth plot. We calculate a translation vector (indicated by the red arrow \red{$\rightarrow$}) that points from the plot center (indicated by $\times$) to the $argmax$ bin (indicated by $\blacksquare$), i.e., the bin with the maximal vote. The center of voting space indicates zero translations and each bin represents a 0.2 $\times$ 0.2 meters region. As a reference, we also provide the ground truth flow (in digits) at the bottom of the ground truth figures. Overall, the $argmax$ bin in the voting space has a similar translation as the ground truth flow in both direction and displacement.

\subsection{Ablation study }
This section studies the impact of the key hyperparameters and the choice of decoder in model design.

\subsubsection{Impact of $M$ and $N$}
\begin{table}
\centering
\resizebox{\columnwidth}{!}{
\begin{tabular}[width=0.45\textwidth]{cccccccc}

\toprule
\multirow{3}{*}{$M$}             & \multirow{3}{*}{$N$}   & \multirow{3}{*}{Latency (ms)}                 & \multicolumn{5}{c}{Bucket Normalized EPE ($\downarrow$)}                         \\ 
\cline{4-8}
                                    &                              &                                   & \multicolumn{5}{c}{Dynamic (normalized EPE)} \\
\cline{4-8}
                                    &                               &                           &\multicolumn{1}{c}{Mean} & Car & O. V. & Pd. & W. V           \\ 
\midrule    
4 & 128 & 27.3$\pm$5.0 & 0.337 & 0.223 & 0.355 & 0.434 & 0.338 \\
8 & 128 & 25.6$\pm$5.2 & \textbf{0.335} & 0.222 & 0.347 & 0.424 & 0.347 \\
16 & 128 & 27.2$\pm$5.2 & 0.337 & 0.223 & \textbf{0.341} & 0.444 & 0.338 \\
32 & 128  & 28.7$\pm$5.2 & 0.336 & 0.222 & 0.350 & 0.434 & 0.337 \\
64 & 128 & 81.9$\pm$17.3 & 0.339 & \textbf{0.221} & 0.352 & 0.442 & 0.341 \\
 \midrule
8 & 64  & \textbf{25.3$\pm$4.3} & 0.343 & \textbf{0.221} & 0.362 & 0.449 & 0.341 \\
8 & 256 & 27.4$\pm$5.4 & 0.337 & 0.222 & 0.370 &\textbf{ 0.422} & \textbf{0.334} \\


\bottomrule
\end{tabular}
}
\caption{\textbf{Ablation study on $M$ and $N$ on Argoverse 2 \underline{val} split.} We empirically test the influence of $M$ and $N$ on VoteFlow. Latency is measured on a single A100 GPU.  }
\label{tab:ablation_mn}
\end{table}

$M$ and $N$ are two key hyperparameters in our Voting Module. 
Ideally, the $M$ neighbors represent points from the same rigid body.
We deploy k-NN search~\cite{ravi2020pytorch3d} to localize the nearest neighbors at time $t$. 
In contrast, $N$ defines the number of potential correspondences at $t+\Delta t$, which is expected to cover the entire search area. 
We employ the \textit{ball query} function~\cite{ravi2020pytorch3d} to search potential neighbors within a specified radius.
Our experiments in \tab{ablation_mn} showcase that changing $M$ has a marginal impact on the performance.
Raising $N$ does not bring performance gain on Cars and Other Vehicles but benefits the result on small-scale objects, such as Pedestrians and Wheeled VRUs.
This enhancement is likely due to the increased coverage of pillars that may contain an object at time $t+\Delta t$. 
\subsubsection{Choice of Decoder}

\tab{ablation_decoder} compares different decoder choices. SeFlow~\cite{zhang2025seflow} employs a decoder with multiple GRU layers and iteratively refines features. 
We take the same design and insert our Voting Module.
Compared to the SeFlow, VoteFlow with GRU decoder yields a substantially better result in the category Pedestrians but worse in Other Vehicles.  
Additionally, using GRU decoders slows down training. 
We further test a simple setup in our model, which consists of four fully connected layers, and achieve significantly improved performance. 
Hence, our model uses the MLP decoder due to its computation efficiency and outstanding performance. 

\subsection{Qualitative evaluation}
In \fig{qualitative_results}, we show qualitative comparisons against SeFlow~\cite{zhang2025seflow}, the best-performing self-supervised baseline.
Colors indicate directions and color saturation indicates the scale of the flow estimation.
In the first row, we show the scene flow of a long bus, highlighted in the red box. Our method generates consistent predictions across nearly the entire vehicle, revealed by uniform coloring and coherent saturation. In contrast, predictions from SeFlow focus primarily on the front section, indicating a lack of rigid prior. 
Nevertheless, our method still encounters difficulties in yielding completely accurate predictions for this larger object.
Similarly, in the second row, the SeFlow predicted flow on the object instance appears to be inconsistently colored compared to the ground truth, whereas our model achieves more uniform coloring overall. 
In the third row, SeFlow generates false positive predictions (FP) over a static object, implying that motion rigidity was not captured for the entire object. Despite a few false negatives, VoteFlow's predictions remain largely consistent with the ground truth. 
Qualitative results show that the architectural inductive bias in our VoteFlow improves the motion rigidity in scene flow prediction.

\begin{table}[!htbp]
\centering
\resizebox{\columnwidth}{!}{
\begin{tabular}[width=0.45\textwidth]{ccccccc}

\toprule
\multirow{3}{*}{Decoder}              & \multirow{3}{*}{\# Params (M)}     & \multicolumn{5}{c}{Bucketed Normalized EPE ($\downarrow$)}                         \\ 
\cline{3-7}
                                        &                               & \multicolumn{5}{c}{Dynamic (normalized EPE)} \\
\cline{3-7}
                                        &                           &\multicolumn{1}{c}{Mean} & Car & O. V. & Pd. & W. V           \\ 
\midrule    
SeFlow ~\cite{zhang2025seflow}           &          0.077 & 0.369 & 0.234 & \textbf{0.342} & 0.541 & 0.358 \\ 

VoteFlow (GRU decoder~\cite{zhang2024deflow})      &   0.100   &    0.354   &  \textbf{0.221} & 0.374 & 0.475  & \textbf{0.344} \\
VoteFlow (MLP decoder) & \textbf{0.087} & \textbf{0.335 }& 0.222 & 0.347 & \textbf{0.424} & 0.347 \\
\bottomrule
\end{tabular}
}
\caption{\textbf{Ablation Study on different decoder choices on Argoverse 2 \underline{val} split.} 
}
\label{tab:ablation_decoder}
\end{table}

%% file: sec/5_conclusions.tex
\section{Conclusions}
\label{sec:conclusions}

Our novel method, VoteFlow, is a state-of-the-art self-supervised scene flow estimation method.
By using a novel Voting Module, we match pillar features across local spatial regions of subsequent LiDAR scans.
This presents a new approach to incorporate motion rigidity inductive bias into scene flow estimation, distinct from loss-based and clustering-based approaches found in prior work.
Our extensive experiments show that VoteFlow is competitive or outperforms other self-supervised models on Argoverse 2 and Waymo,
and also when testing on Waymo \textit{even without training on Waymo}.

\paragraph{Limitations and future work.} 
In our work, the size of the voting space $V$, $H_v \times W_v$, depends on the magnitude of potential translation $x_{max}$ and $y_{max}$. As $\Delta t$  increases, the maximum possible translation grows, expanding the voting space and increasing the computational burden. 
In future work, we aim to extend the voting strategy to incorporate rotations alongside translations and enhance the voting module's efficiency, particularly for longer time intervals. 

Furthermore, fusing multi-modal information of an autonomous driving car for joint scene flow and optical flow estimation is also of our interest~\cite{wang2024unibev}.

%% file: sec/6_acknowledgement.tex
\newline

\noindent \textbf{Acknowledgement}. 
Y. Lin was supported by NWO NGF-AiNed XS (file number: NGF.1609.23.015). S. Wang was supported by the 3D Urban Understanding (3DUU) Lab funded by the TU Delft AI Initiative.

%% file: sec/X_suppl.tex
\clearpage
\setcounter{page}{1}
\maketitlesupplementary
\renewcommand{\thetable}{\Roman{table}}

\section{Ablation studies on ground removal}
\label{supp:ground_removal}
We followed the official Argoverse 2 benchmark and removed the ground with rasterized maps in VoteFlow. \tab{ablation_gm} now includes performance after ground removal using Patchwork++ \cite{lee2022patchwork++}, an offline algorithm.
Without maps, performance drops notably, particularly on pedestrians (Pd.) and wheeled VRUs (W.V).
Compared to rasterized maps, Patchwork++ achieves $95.5\%$ precision and $82.3\%$ recall in ground classification.
Cases without flat ground are not well studied in current benchmarks, which remains an open challenge.

\begin{table}[!h]
\centering
\resizebox{1.0\columnwidth}{!}
{
\begin{tabular}[width=0.5\textwidth]{ccccccc}

\toprule

Ground removal                       & Mean & Car & O. V. & Pd. & W. V           \\ 
\midrule    
Patchwork++    & 0.369 & 0.246 & 0.364 & 0.477 & 0.389 \\ 
Rasterized maps  & \textbf{0.335} & \textbf{0.222} & \textbf{0.347} & \textbf{0.424} & \textbf{0.347} \\
\bottomrule
\end{tabular}
}
\caption{\textbf{Impact of ground removal on Argoverse 2 \underline{val} split.}
}
\label{tab:ablation_gm}
\end{table}

\section{Ablation studies on the pillar size}
\label{supp:pillar_size}

\tab{ablation_pillar_size} shows decreasing pillar size improves accuracy but increases latency. Our default size $0.2m$ balances performance and latency (on an A100 GPU).

\begin{table}[h]
\centering
\resizebox{0.95\columnwidth}{!}{
\begin{tabular}[width=1.0\textwidth]{cccccccc}
\toprule
Pillar (m)  & Latency (ms)  & Mean   & Car   & O.V.   & Pd.  & W. V.  \\
\midrule    

$0.1$  &  58.6$\pm$6.5  & \textbf{0.327} & \textbf{0.216} & 0.360 & \textbf{0.403} & \textbf{0.331} \\ 
$0.2$  & 25.6$\pm$5.2 & 0.335 & 0.222 &\textbf{0.347}& 0.424 & 0.347 \\
$0.4$  &  \textbf{17.1$\pm$4.6} & 0.371 & 0.226 & 0.369 & 0.524 & 0.366 \\ 
\bottomrule
\end{tabular}
}
\caption{\textbf{Ablation study on pillar sizes on Argoverse 2 \underline{val} split.}  
}
\label{tab:ablation_pillar_size}
\end{table}

\section{Ablation studies on loss functions}
\label{supp:loss_ablation}

\begin{table}[h]
\centering
\resizebox{\columnwidth}{!}{
\begin{tabular}[width=0.45\textwidth]{ccccccc}

\toprule
\multirow{3}{*}{Method}        & \multirow{3}{*}{Loss}                      & \multicolumn{5}{c}{Bucket Normalized EPE ($\downarrow$)}               \\ 
\cline{3-7}
                                &                                   & \multicolumn{5}{c}{Dynamic (normalized EPE)}             \\
\cline{3-7}
                                &                                &\multicolumn{1}{c}{Mean} & Car & O. V. & Pd. & W. V           \\ 
\midrule    
 SeFlow~\cite{zhang2025seflow}                       &  $\mathcal{L}_{total}$        &         0.369  &  0.234  & \textbf{0.342} & 0.541 & 0.358   \\
VoteFlow (Ours)                 &  $\mathcal{L}_{total}$        &       \textbf{0.335}  &  \textbf{0.222}  & 0.347 & \textbf{0.424} & \textbf{0.347}   \\ 
 \midrule
 SeFlow \cite{zhang2025seflow}  & $\mathcal{L}_{chamfer}$       &            0.463 & 0.347 &  0.579 &  0.541 & 0.386 \\
 VoteFlow (Ours)                & $\mathcal{L}_{chamfer}$       &            \textbf{0.444}  &  \textbf{0.320}  & \textbf{0.563} & \textbf{0.511} & \textbf{0.381}   \\ 
\bottomrule
\end{tabular}
}
\caption{\textbf{Impact of loss functions.} 
All results are from Argoverse 2 \underline{val} split.
We test the performance of our model and the baseline SeFlow~\cite{zhang2025seflow}, which have been trained by only $\mathcal{L}_{chamfer}$. This makes sure the model has not been regularized by any other loss functions related to motion rigidity.  
The performance improvement over SeFlow on Dynamic Mean indicates the benefit of the Voting Module in our design.
}
\label{tab:ablation_loss}
\end{table}

SeFlow~\cite{zhang2025seflow} employs multiple losses to enforce consistent flow predictions. For example, the cluster loss $\mathcal{L}_{cluster}$ encourages consistent flow prediction from the same cluster; $\mathcal{L}_{static}$ directly forces the static flows to be zeros; $\mathcal{L}_{dynamic}$ is explicitly imposed on points that are classified as dynamic in preprocessing. 
The usage of these losses shares a purpose similar to our Voting Module, i.e., to make flow prediction consistent. 
To evaluate the impact of our Voting Module in an isolated environment, we conduct ablation studies where no additional losses are adopted other than $\mathcal{L}_{chamfer}$.
\tab{ablation_loss} compares the performance of SeFlow and VoteFlow with $\mathcal{L}_{total}$ and $\mathcal{L}_{chamfer}$ alone (excluding $\mathcal{L}_{cluster}$, $\mathcal{L}_{dynamic}$, and $\mathcal{L}_{static}$). 
Both models show a significant performance drop when trained only with $\mathcal{L}_{chamfer}$, indicating the importance of the explicit loss as regularization. 
However, VoteFlow still outperforms SeFlow across all categories, with a notable $2.1 \%$pt improvement in Dynamic Mean, demonstrating the effectiveness of the Voting Module. 

\begin{figure*}[h!]
    \centering
    \includegraphics[width=\textwidth]{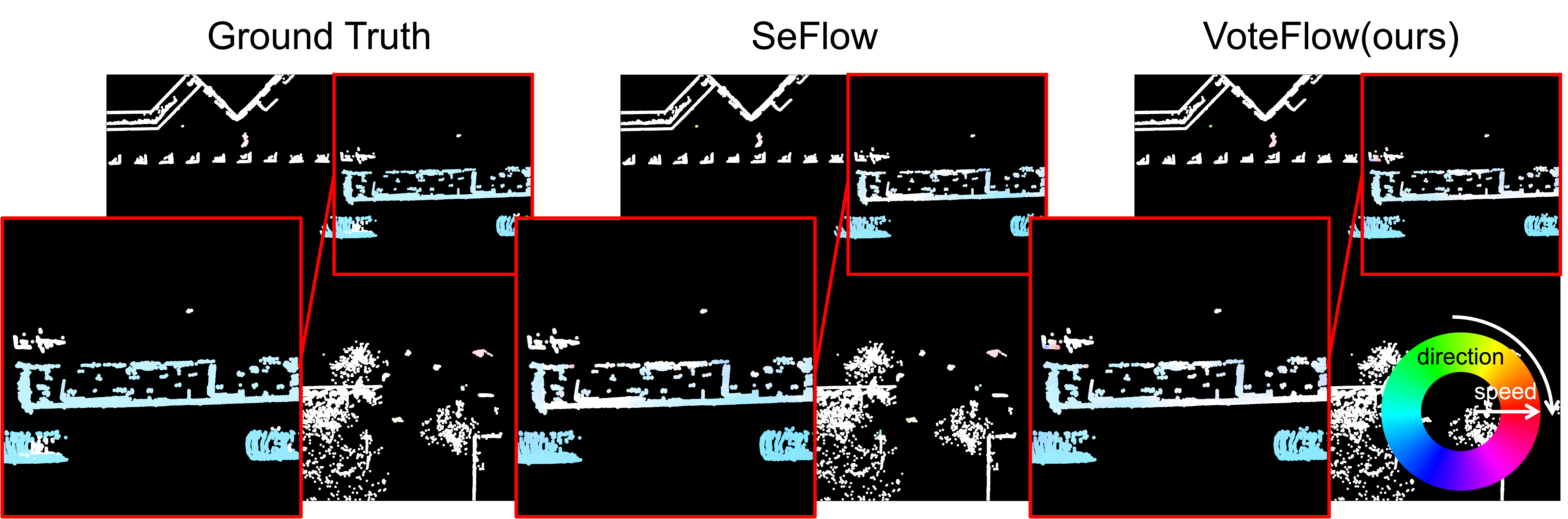}
    \caption{\textbf{Failure cases on Argoverse 2 validation set.} Colors indicate directions and saturation of the color indicates the scale of the flow estimation. Both SeFlow and VoteFlow struggle to predict consistent flows for large-size, rigidly moving objects.
}
    \label{fig:failure_qualitative}
\end{figure*}

\section{Ablation studies on the Voting Module and voting features}
\label{supp::voting_ablation}

\begin{table}[!h]
\centering
\resizebox{\columnwidth}{!}
{
\begin{tabular}[width=0.5\textwidth]{cccccccc}
\toprule
Voting             & Voting Feats         & Mean & Car & O. V. & Pd. & W. V           \\ 
\midrule    

 \xmark    &   \xmark & 0.373  & 0.222  & 0.397 & 0.512 & 0.362 \\ 
 
$\checkmark$  & Fused feats \textbf{G} & 0.348 & \textbf{0.220} & 0.383 & 0.445 & \textbf{0.344} \\
\midrule
$\checkmark$  & Pillar feats \textbf{I}  & \textbf{0.335} & 0.222 & \textbf{0.347} & \textbf{0.424} & 0.347 \\
\bottomrule
\end{tabular}
}
\caption{\textbf{Ablation study on the Voting Module and voting features.} Taking (separated) features from the Pillar Feature Net further enhances performance over taking fused features from the U-Net.}
\label{tab:ablation_feats}
\end{table}

\tab{ablation_feats} firstly compares models of the same architecture without (\xmark) and with ($\checkmark$) the Voting Module. 
Adding voting improves the mean by $3.8\%$pt. 
Additionally, we explored multiple configurations for the input features of our Voting Module.
In the first configuration, we use the separate pillar indices $\textbf{P}^{t}$ and $\textbf{P}^{t+\Delta t}$ from the input point clouds to retrieve per-point features from fused feature $\mathbf{G}$.
This design outperforms its counterpart without the Voting Module by a margin of $2.5\%$pt in Mean Dynamic, indicating $\mathbf{G}$ effectively encodes the fused semantics from both point clouds.
An alternative design feeds the pseudo images $\mathbf{I^{t}}$ and $\mathbf{I^{t+\Delta t}}$ directly into the Voting Module.
In contrast to $\mathbf{G}$, $\mathbf{I}$s are separate pillar features for source and target point clouds. As shown in \tab{ablation_feats}, the second design achieves further performance improvement over the first one.
We therefore argue that our proposed Voting Module is a universal method for exploiting the motion rigidity prior regardless of which type of features are used as input.

\section{Additional Qualitative Results}
We show a failure case of our model in \fig{failure_qualitative} where it fails to produce consistent flow for larger objects that move rigidly. We suspect the failure originates from the fact that the predefined local neighborhood where local rigidity holds is not large enough to cover the entire object.
Future work could explore adjusting the neighborhood range adaptively for such cases.